\documentclass[journal]{IEEEtran}

\usepackage{cite}
\usepackage{cuted}
\usepackage{capt-of}
\usepackage{amssymb}
\usepackage{latexsym}
\usepackage{algpseudocode}
\usepackage{blindtext}
\usepackage{algorithm}
\usepackage{times}
\usepackage{epsfig}
\usepackage{amsmath}
\usepackage{textcase}
\usepackage{comment}
\usepackage{breqn}
\usepackage{xcolor,balance}
\usepackage{ragged2e}
\usepackage[normalem]{ulem}
\usepackage{caption, booktabs}

\ifCLASSINFOpdf
  \usepackage{graphicx}
\else
\fi

\begin{document}

\title{Detection and Localization of Firearm Carriers in Complex Scenes for Improved Safety Measures}

\author{Arif~Mahmood$^*$, Abdul~Basit,
        M.~Akhtar Munir,
        Mohsen~Ali%
\thanks{A. Mahmood and A. Basit are with the Center for Artificial Intelligence and Robot Vision (CAIRV), Department
of Computer Science, Information Technology University (ITU), Lahore, Pakistan. M. A. Munir and M. Ali are with the Intelligent Machines Lab (IML), Department
of Computer Science, Information Technology University (ITU), Lahore, Pakistan. 
$^*$ A. Mahmood is the corresponding author. 
E-mails: \{abdul.basit, akhtar.munir, mohsen.ali, arif.mahmood\}@itu.edu.pk}}
\maketitle
\begin{strip}
    \centering
    \includegraphics[width=0.95\linewidth, keepaspectratio]{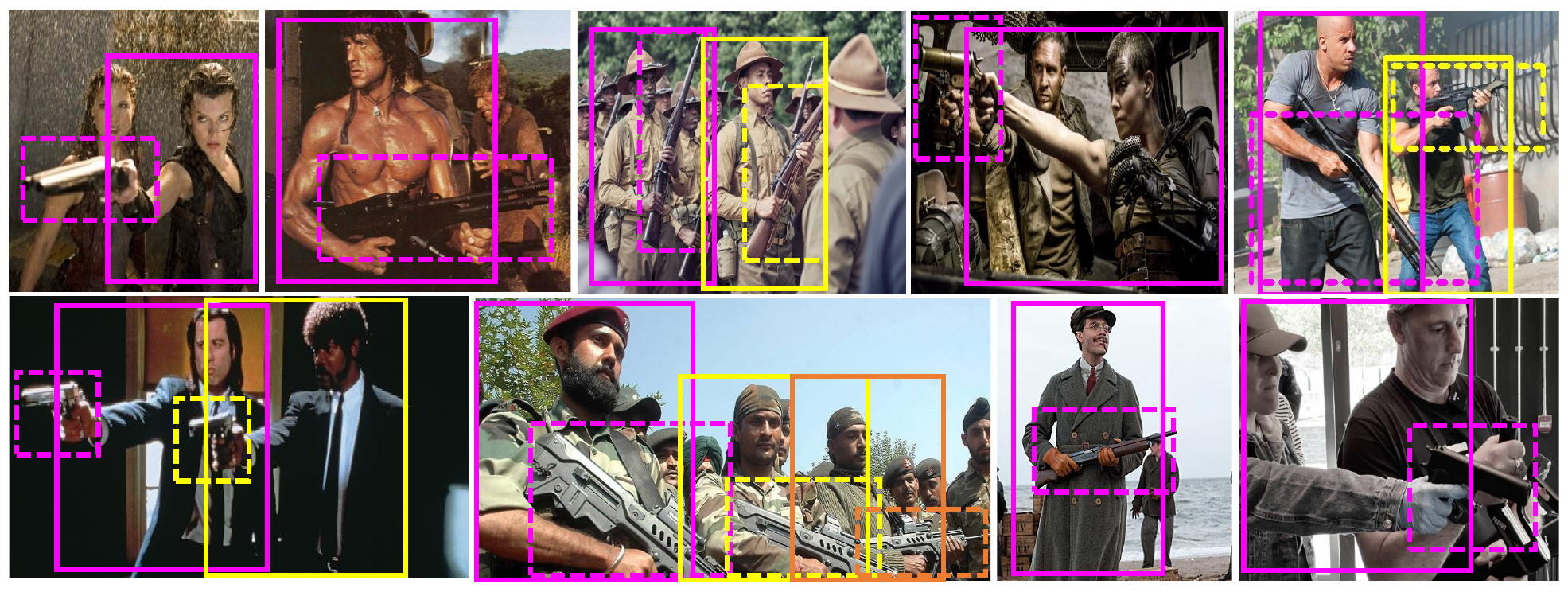}
    \captionof{figure}{The proposed algorithm associates a firearm with its carrier. Full lines show the human detections, and dotted lines with the same color show the associated firearms as detected by the proposed algorithm.}
    \label{fig:teaser}
\end{strip}

\begin{abstract}

Detecting firearms and accurately localizing individuals carrying them in images or videos is of paramount importance in security, surveillance, and content customization. However, this task presents significant challenges in complex environments due to clutter and the diverse shapes of firearms. To address this problem, we propose a novel approach that leverages human-firearm interaction information, which provides valuable clues for localizing firearm carriers. Our approach incorporates an attention mechanism that effectively distinguishes humans and firearms from the background by focusing on relevant areas. Additionally, we introduce a saliency-driven locality-preserving constraint to learn essential features while preserving foreground information in the input image. By combining these components, our approach achieves exceptional results on a newly proposed dataset. To handle inputs of varying sizes, we pass paired human-firearm instances with attention masks as channels through a deep network for feature computation, utilizing an adaptive average pooling layer. We extensively evaluate our approach against existing methods in human-object interaction detection and achieve significant results (AP=77.8\%) compared to the baseline approach (AP=63.1\%). This demonstrates the effectiveness of leveraging attention mechanisms and saliency-driven locality preservation for accurate human-firearm interaction detection. Our findings contribute to advancing the fields of security and surveillance, enabling more efficient firearm localization and identification in diverse scenarios.

\end{abstract}


\IEEEpeerreviewmaketitle
\vspace{-2mm}
\section{Introduction}
\label{sect:intro}


\IEEEPARstart{F}{irearms}-related violence remains an inadequately researched area, despite the alarming increase in the annual death toll worldwide \cite{TCSS1,newyear2020, Americas23:bbc, amnesty}. To address this pressing issue, many governmental and private entities have deployed surveillance systems in public areas such as hospitals, universities, parks, and malls. However, these systems often require a significant number of personnel to manually monitor the captured footage, which presents several challenges. The reliance on human observation for extended periods of time poses a risk of errors, as even a momentary lapse in concentration  can lead to disastrous consequences \cite{howard2013suspiciousness,TCSS2}. Therefore, there is a critical need for an innovative and reliable method capable of automatically detecting firearms and simultaneously highlighting the individuals carrying them. This would enable immediate actions by the relevant authorities \cite{TCSS4}. Integrating such a system into monitoring devices would significantly enhance the identification of potential threats in dynamic environments \cite{TCSS3}. Furthermore, this solution could assist in filtering multimedia content, based on age and target audience, that contains gun violence or any elements associated with firearms.


Detecting firearms presents a formidable challenge due to the wide range of variations in color, appearance, and texture that firearms exhibit worldwide. Moreover, each firearm possesses a distinct shape and size, ranging from compact handguns to lengthy rifles, further exacerbating the complexity of the problem. In this paper, we specifically address the intricate task of localizing firearm carriers, which becomes even more challenging due to the various configurations in which firearms can interact or coexist with individuals. These scenarios highlight the difficulties encountered, such as the potential for small-sized guns to be easily overlooked amidst clutter and occlusion. Additionally, large-sized, slender rifles can span across multiple human carriers, with overlapping features further complicating the problem. Consequently, discerning which specific individual is carrying or holding the firearm becomes an intricate endeavor requiring sophisticated approaches.


Existing firearms detection frameworks often fail to identify firearm carriers, leaving a critical gap in the overall detection process \cite{oaod}. To overcome this limitation, we propose a novel approach that incorporates a human object interaction mechanism specifically designed for the identification of firearm carriers. Our method focuses on two firearm categories, namely guns and rifles, and takes into account multi-sized instances of human-firearm pairs, considering both interaction and non-interaction scenarios.

{Building upon the preliminary version published in a conference \cite{iciploc}, our current method represents an extension of our previous work. In our earlier study, the interaction between humans and firearms was determined solely based on extended bounding boxes encompassing human-firearm pairs. However, upon analyzing this approach, we discovered that the reliance solely on extended boxes led to an inclusion of a significant portion of background regions along with the foreground, adversely affecting the computation of relevant features. Consequently, this resulted in a degradation of performance.} To address this challenge, our current approach introduces novel strategies to enhance feature computation and mitigate the inclusion of irrelevant background regions. By doing so, we aim to significantly improve the accuracy and robustness of firearm carrier identification in complex scenes.


Our proposed approach incorporates binary attention channels for each object and human, enabling the classifier to assign higher weights to the salient regions associated with the foreground. To construct human-firearm pairs, we consider each human and firearm present in the image, labeling them with interaction or non-interaction categories, and concatenate them with the attention mechanism. Additionally, our architecture integrates a saliency-driven locality-preserving branch that effectively preserves the spatial location and size information of the objects within the input.
This branch utilizes a reconstruction network to enforce the feature extractors to maintain the integrity of the relevant object and human features within the extended bounding box. Figure \ref{fig:teaser} illustrates the output predicted by our proposed method, demonstrating its effectiveness. Throughout this article, the terms ``hold'' and ``carry'' are used interchangeably. For a comprehensive understanding of our approach, the detailed architecture is presented in Figure \ref{fig:main} and thoroughly explained in Section \ref{sect:EHFPL}.


{Our proposed system holds promising potential for integration with smart surveillance systems to enhance firearm carrier detection capabilities. To train the proposed architecture and to facilitate research in this direction, a novel dataset is introduced  providing a collection of images each containing at least one human and one firearm. Each image in the dataset is meticulously annotated with bounding boxes for humans and firearms (including guns and rifles), as well as labels indicating the interaction or non-interaction between them.}

The main contributions of our work are as follows:
\vspace{0.4cm}
\begin{itemize}
    \item We propose a novel deep learning architecture that effectively identifies localized firearm carriers by incorporating human-firearm instance pair detection. This approach enables precise localization and identification of individuals carrying firearms.
    
    \item Our architecture integrates an attention-based mechanism that focuses on relevant regions of interest associated with humans and firearms. By doing so, the model is compelled to allocate greater attention to the queried firearms, improving detection accuracy and robustness.
    \item We introduce a reconstruction loss that promotes saliency-driven locality preservation of relevant objects within the input image. This ensures that essential features of the identified objects are accurately retained while maintaining spatial information.
    \item {We present a new dataset specifically designed for firearm and its carrier identification and localization, consisting of 3,128 images. Each image within the dataset contains at least one firearm and one human. Manual annotations have been meticulously created to label the interacting and non-interacting firearm-human pairs, facilitating comprehensive training and evaluation of our proposed approach.}
    \item We conduct extensive experiments to evaluate the performance of our proposed algorithm. The results demonstrate the superior performance of our method compared to existing approaches, showcasing its effectiveness in firearm carrier detection and localization.
\end{itemize}
\vspace{0.1cm}


Our contributions represent a significant advancement in the field of firearm detection and localization, offering a robust and accurate solution that has the potential to greatly enhance security measures and surveillance systems across a wide range of real-world scenarios. To thoroughly evaluate the effectiveness of our proposed method, we conducted a comprehensive analysis by comparing it against baseline approaches as well as three existing human-object interaction (HOI)-based approaches.

Despite its simple design, which focuses on the classification of human-firearm pairs, our method exhibits remarkable stability and delivers excellent performance. This demonstrates the effectiveness and efficiency of our approach, which outperforms the baseline methods and achieves comparable or even superior results to the state-of-the-art HOI-based approaches.

Our findings underscore the significance of leveraging the specific problem of human-firearm interaction, enabling us to develop a highly effective solution without the need for overly complex architectures. This simplicity not only enhances the practicality and ease of implementation of our method but also ensures its suitability for real-time applications where computational efficiency and responsiveness are critical factors.

In summary, our contributions significantly advance the field of firearm detection and localization, providing a robust and accurate solution that can effectively enhance security measures and surveillance systems in a wide range of real-world scenarios. Through comprehensive comparisons with existing approaches, our method demonstrates stability, excellent performance, and the potential to address the challenges associated with classifying human-firearm pairs in a simplified yet effective manner.

\vspace{-2mm}
\section{Related Work}
\subsection{Object Detection:}

Numerous algorithms have been proposed that utilize convolutional neural networks (CNNs) for object detection in images \cite{ren2015faster,redmon2018yolov3,liu2016ssd,lin2017feature, tian2019fcos, duan2019centernet}. These object detection algorithms can be broadly categorized into two groups: two-stage detectors and one-stage detectors. Two-stage detectors \cite{ren2015faster, singh2018analysis, lin2017feature} employ a Region Proposal Network (RPN) that extracts object regions in a class-agnostic manner using anchor box information. This approach involves a two-step process of generating region proposals followed by classification and refinement.

On the other hand, one-stage object detectors \cite{redmon2018yolov3,redmon2017yolo9000,redmon2016you,liu2016ssd, fu2017dssd, tian2019fcos,duan2019centernet} offer a trade-off between speed and accuracy, as they are designed to be faster while still achieving reasonable detection performance. Although some of these detectors utilize default boxes as anchor boxes, they do not incorporate a separate RPN in their pipeline. By leveraging the capabilities of CNNs, both two-stage and one-stage detectors have significantly advanced the field of object detection, providing efficient and effective solutions for various applications. The choice between these two categories depends on the specific requirements of the task at hand, considering factors such as accuracy, speed, and computational resources.


\begin{figure*}[t]
    \begin{center}
    \includegraphics[width=1.0\linewidth]{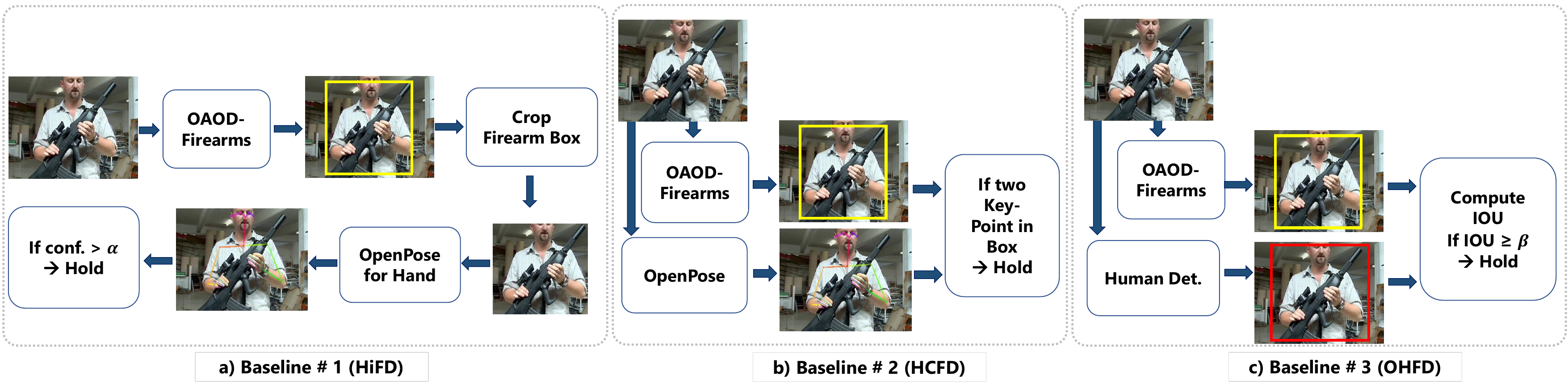}
    \caption{Baseline Approaches: a) Hands in Firearm Detections (HiFD): detected firearm bounding box is input to a hand pose detector.
    b) Hand-pose Conditioned Firearm carrier Detection (HCFD): Firearm and hand key points are detected in parallel on full image. 
    c) Overlap  of  Human  and  Firearms  Detections (OHFD): Full human and firearm are detected in parallel and intersection over union (IoU) is computed.
    }
    \label{fig:base}
    \end{center}
\end{figure*}


\subsection{Firearm Detection:}
\label{sec:fd}

Firearms exhibit diverse shapes, sizes, appearances, and textures, making their accurate detection a challenging task for generic object detectors. Unfortunately, despite the significance of this area, the number of dedicated firearm detectors remains limited. Javed et al. \cite{oaod,iqbal2019orientation} proposed a weakly supervised system called the Orientation Aware Object (firearm) Detection (OAOD) system, which utilizes orientation information and axis-aligned bounding boxes to extract prominent firearm regions from the background. This framework demonstrates strong performance in firearm detection.

Olmos et al. \cite{olmos2018automatic} employed Faster R-CNN \cite{ren2015faster} for gun detection. However, their system focuses exclusively on hand-guns, representing only a subset of firearms. Akcay et al. conducted gun detection in x-ray baggage security imagery using various object detection strategies, including both two-stage and one-stage detectors \cite{akcay2018using}. Their work primarily targeted firearms concealed in baggage screening. {Rogers et al.~proposed a method for detecting Small Metallic Threats (SMTs) hidden among legitimate goods using X-ray cargo imagery \cite{Rogers2017ADL}. While not specifically focused on firearms, their approach addresses the broader challenge of detecting concealed threats.}

Although firearm detection has attracted some research interest in recent years, no previous work has addressed the identification of firearm carriers in an image. In this study, we present an automated system that formulates human-firearm pairs to answer the question of who is carrying the firearm, contributing as the first attempt to address this specific aspect of firearm detection.

\subsection{Human Object Interaction:}

Human-Object Interaction (HOI) has recently emerged as a prominent topic in the field of deep learning and computer vision. In our study, we formulate our problem by leveraging HOI-based strategies, specifically focusing on the interaction and non-interaction of objects with humans. HOI approaches often rely on contextual information within the image to determine the interaction between humans and objects. For instance, Wang et al. \cite{wang2019deep} utilize interaction points to localize and classify interactions between human-object pairs. Exploiting pose and action information has also proven effective in accurately identifying human-object interactions \cite{gkioxari2018detecting}. Various pose extraction techniques, such as Hand, OpenPose, CPM, and CrowdPose \cite{simon2017hand, cao2018openpose, wei2016cpm, li2019crowdpose}, have been employed to learn human-object interactions in images. Gao et al. \cite{gao2018ican} propose a trainable instance-centric attention module that utilizes appearance features of instances to capture interactiveness. Some approaches, like that of Li et al. \cite{li2019transferable}, transfer interactiveness to perform multitask learning, combining common schemes used by HOI-based methods. 

Li et al. \cite{li2020pastanet} introduce part states annotations and utilize part states (PaSta) for HOI classification to improve the final output. Their hierarchical graph-based model determines which part focuses on and contributes to the final prediction. However, if part annotations are unavailable in the datasets, the PaSta approach cannot be applied. Liao et al. \cite{liao2020ppdm} propose an end-to-end single-stage architecture that detects humans, objects, and interactions simultaneously. {While some recent methods \cite{lim2023ernet,park2023viplo,ma2023fgahoi} achieve higher performance by employing complex architectures with a significantly increased number of parameters, it is worth noting that such approaches may be impractical for low-cost real-time systems in the field.}

In our paper, we incorporate strategies and knowledge from object detection, pose estimation, and HOI detection to propose baseline methods for detecting and localizing firearm carriers in images. Specifically, we introduce three baseline methods that utilize human and hand pose detection followed by firearm detection to identify firearm carriers in complex scenes where multiple humans and multiple firearms are present. Based on an analysis of these baseline approaches and \cite{iciploc}, we propose a novel architecture that incorporates attention mechanisms and saliency-driven reconstruction to accurately identify HOIs in images.

\section{Baseline Approaches}
\label{baselines}
\vspace{-0.2mm}
\subsection{Detecting Hands inside Firearm Detections (HiFD)}
\label{sect:HiFD}

In this approach, we utilize the OAOD algorithm \cite{oaod} to detect firearms. We then employ a localized firearm bounding box as a probable hand location and input these hand regions into the Multi-view Bootstrapping algorithm \cite{simon2017hand}, which returns a set of hand keypoints. To filter out keypoints with low confidence, we introduce a threshold parameter, denoted as $\alpha$. If any two keypoints have confidence scores greater than the threshold (in our case, set to 0.3), we classify the detected firearm as being carried by a person.

This approach demonstrates relatively higher accuracy for small-sized firearms, such as guns (as shown in Table \ref{tab:clsgroundtruth}). However, the accuracy decreases when the size of the axis-aligned bounding box of the detected object is larger, as is the case with long rifles. The main reason for this performance decrease is that the larger bounding boxes are not centered around the hands, and the presence of background clutter, including hands, within the larger bounding box area leads to failures in keypoint estimation.

\vspace{-4mm}
\subsection{Pose Conditioned Firearm carried Detection (HCFD)}
\label{sect:HCFD}

In addition to performing firearm detection using \cite{oaod}, we utilize OpenPose \cite{cao2018openpose} to estimate key points of the entire human body. However, for the purpose of inspecting firearm bounding boxes, we focus solely on the hand key points. In this baseline approach, we classify a firearm as ``carried'' if at least two of the estimated hand key points are located inside the detected firearm bounding box. Otherwise, it is categorized as ``not-carried''.

One limitation of this approach is that the full-body pose, particularly the hand key points, may not always be visible due to occlusions or partial appearance. If the elbow or wrist, which are used to determine the probable hand location, are occluded (which is often the case), the hand key points cannot be accurately detected. Consequently, the performance of this approach is significantly affected, as it heavily relies on the performance of the pose detector.

\subsection{Overlap of Human and Firearms Detections (OHFD)}
\label{sect:OHFD}

In this approach, human detection is performed using Faster RCNN \cite{ren2015faster} with ResNet-101, which is pretrained on MS-COCO dataset. For firearm detection, we utilize the OAOD algorithm \cite{oaod}. To determine whether a firearm is being carried by a human, we compute the Intersection over Union (IoU) between all detected firearms and human bounding boxes. If the IoU between a firearm and a human is the highest among all other humans and exceeds a threshold of $\beta$ (where $\beta = 0.5$), the firearm is considered to be carried by that human. However, associations with an IoU lower than 0.5 are discarded. It is important to note that this approach may encounter challenges in complex scenes where firearm bounding boxes may have a larger overlap with non-carriers, leading to potential misclassifications and decreased performance.
\begin{figure*}[t]
    \begin{center}
    \includegraphics[width=1.0\linewidth]{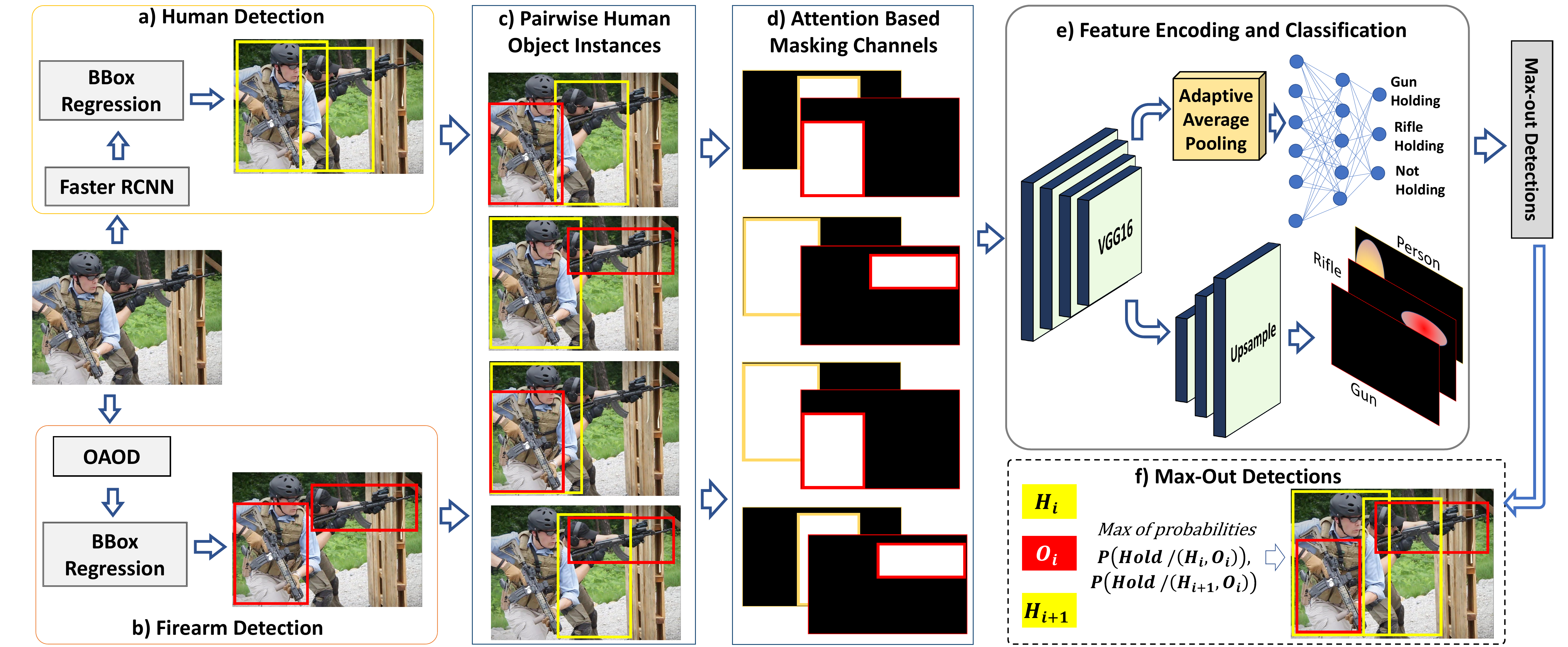}
    \caption{Framework of the proposed firearm carrier detection algorithm:
    (a) Human detection by Faster RCNN
    (b) Detection of firearms by OAOD 
    (c) Formation of all possible human-object instance associations.
    (d) Generation of attention channels using  detectors' output
    (e) Concatenation of attention channels with appearance channels and input to the network, which classify each human-object instance association along with the reconstruction network.
    (f) Max-out detection is used to resolve multiple associations of humans and objects, which results in final detection.
    }
    \label{fig:main}
    \end{center}
\end{figure*}

\section{Proposed  Human Firearm Pair Localization Algorithm}
\label{sect:EHFPL}
The first two baseline approaches in  Sec.~\ref{baselines}  only  classify a firearm  being carried or not carried. In the current work we also aim to localize the carrier of the firearm.
These approaches suffer performance degradation because of the challenges in human pose estimation. 
The third approach in Sec. \ref{sect:OHFD}  fails to achieve good performance in complex scenes, where the main body of the firearm carrier is not necessarily close to the firearm. Therefore it is important to learn features that associate a firearm with its human carrier. 

\subsection{Preliminaries}
\label{sect:prelim-HFPL}

The main architecture of our proposed approach is illustrated in Fig. \ref{fig:main}.
Let $I \in \mathbf{R}^{w \times l \times 3}$ denote the input image, where $w$ and $l$ represent the width and height of the image, respectively.
We denote the bounding box of the $h$-th human as $b_h$, and the bounding box of the $o$-th firearm as $b_o$. Here, $h$ ranges from 1 to $n_h$, and $o$ ranges from 1 to $n_o$, where $n_h$ and $n_o$ denote the total numbers of humans and firearms in the image $I$, respectively.
Our goal is to identify human-firearm pairs by predicting the bounding boxes $\hat{b}_h$ and $\hat{b}_o$ with high probabilities $p_h$ and $p_o$, respectively, and infer the probability $p_{o,h}$ of their carrying relationship. A human-firearm pair is considered ``carried'' if the person in the image, identified by bounding box $\hat{b}_h$, is carrying a firearm with bounding box $\hat{b}_o$.

In all the algorithms, we rely on off-the-shelf models to predict the bounding boxes of humans and firearms, as well as their corresponding probabilities.
Let $\hat{n}_o$ denote the number of detected firearm bounding boxes, and $\hat{n}_h$ denote the total number of detected humans.
For each human, denoted by $\hat{b}_h$, predicted with confidence $\hat{p}_h$ by the off-the-shelf human detector, we infer the bounding box $\hat{b}_o$ and the class of the firearm prediction $\hat{c}$ with confidence $\hat{p}_o$.
To process the pairs, we create a paired bounding box (PBB) by taking the union of the inferred bounding boxes of the human $h$ and firearm $o$, forming the smallest bounding box that contains both.
For each PBB, we crop the corresponding region from the input image to generate pair instances.
Given $\hat{n}_o$ detected objects and $\hat{n}_h$ detected persons, the total number of PBBs is $n = \hat{n}_o \times \hat{n}_h$.
Some of these paired bounding boxes represent associated pairs (carrying or holding), while others may not be associated.
The proposed approach will be discussed in detail in the following sections.

\subsection{Firearm and Human Detection}
\label{sect:det-EHFPL}

We utilize the Orientation Aware Object Detector (OAOD) \cite{oaod}, an advanced algorithm renowned for its effectiveness in firearm detection, to identify firearm instances within an image (for more information, refer to Sec. \ref{sec:fd}). Simultaneously, we employ Faster RCNN with a ResNet-101 backbone, trained on the MS-COCO dataset, to detect humans. For training purposes, we utilize the ground truth bounding boxes available in the training dataset.

\subsection{Pairwise Human Firearm Instances}
\label{sect:inst-EHFPL}

Using the predicted bounding boxes for humans and firearms, we construct what we refer to as a \textit{Paired Bounding Box} (PBB) that encompasses both the firearm and the corresponding human. The subsequent stage of our architecture is trained to discern whether these PBBs depict an interaction or not. To train the classifiers, we manually assign interaction and non-interaction labels to these paired ground truth bounding boxes. By comparing the PBBs generated from the training data with the ground truth, we ascertain the accuracy of their predicted labels. Each PBB represents a cropped region containing both the human and firearm instances.

\subsection{Attention Based Masking Channels}
\label{sect:atten-EHFPL}

Our base algorithm directly utilize the Paired Bounding Box (PBB) for classification. However, it often contains irrelevant information that can have a detrimental impact on overall performance. In our current work, we enhance the PBB representation by incorporating additional information in the form of attention channels, as illustrated in Fig. \ref{fig:main}. Let $I \in \mathcal{R}^{w\times h\times K}$ denote a PBB, where $w$ and $h$ represent the width and height, and $K$ denotes the number of channels. For RGB or YCbCr images, $K=3$, while for grayscale images, $K=1$. We further process the PBB by including three attention channels: gun attention mask $\text{M}{g}$, rifle attention mask $\text{M}{r}$, and human attention mask $\text{M}{h}$. These attention channels are concatenated with the original image, resulting in $I \oplus {M}{g} \oplus {M}{r} \oplus {M}{h}$. In the case of RGB or YCbCr images, the number of channels increases to $K=6$, and for gray-scale images, it becomes $K=4$. By incorporating attention masks, we guide the network to focus on the primary objects relevant to the interaction, thereby improving the efficiency of human-object interaction identification.

\subsection{Feature Encoding and Classification}
\label{sect:fcn-EHFPL}
The feature extraction and classification framework is shown in Fig. \ref{fig:main}(e).
It includes the following main components.

\subsubsection{Feature Encoding Block}

The cropped PBBs are concatenated with the Attention Based Masking Channels to create the Attention Augmented PBB (\textit{aPBB}), which is then fed into the Feature Encoding Block (FEB) for feature extraction. The FEB consists of a convolutional neural network responsible for extracting the feature volume. To handle the varying spatial dimensions of the \textit{aPBB}, an \textit{Adaptive Average Pooling} layer is applied. In our experiments, we utilize a pre-trained VGG16 network for feature extraction. By default, the VGG16 network expects input with 3 channels representing RGB values. However, in our case, the \textit{aPBB} can have either 4 or 6 channels ($K = {4, 6}$) depending on the color space of the image. To accommodate this, the first convolutional layer of the VGG16 network is replaced with a new convolutional layer that accepts an input size of 4 or 6 channels. These newly introduced layers are randomly initialized before training, while the other layers utilize pre-trained weights from VGG16 trained on ImageNet \cite{imagenet_cvpr09}.

The feature volume extracted from the convolutional layers has a size of $f_e \in \mathcal{R}^{w' \times h' \times 512}$, where $w'$ and $h'$ vary depending on the size of the PBBs. To ensure compatibility with the required output dimensions, \textit{Adaptive Average Pooling} (AAP) is applied, which performs a resize operation using estimated kernel sizes. The feature volume from VGG16 is resized to $fb \in \mathcal{R}^{7 \times 7 \times 512}$ using AAP. Besides achieving a predefined size, AAP also extracts discriminative information, enhances the expressiveness of the features, and enables the capture of global morphological details of the primary object.

\subsubsection{PBB Classification Module}

The compact output features extracted from AAP are then passed to the PBB classification network. This network is trained to classify the \textit{aPBBs} into three classes: gun-human, rifle-human, and no interaction. For the training of this network, the PBBs are manually annotated. The PBB classification network consists of three fully connected layers followed by a softmax layer. To train both the PBB classification network and FEB, we utilize the cross-entropy loss function defined as follows:
\begin{equation}
\label{eq:closs}
\centering
\mathcal{L}_c(p_c, g_c) = \sum_{i=1}^{n_{s}}\sum_{j=1}^{n_c} g_c(i,j) \log(p_c(i,j)),
\end{equation}
were, $p_c \in \mathcal{R}^{n_c}$ represents the predicted class probabilities for gun-human, rifle-human, and no interaction, while $g_c$ is a one-hot encoded vector representing the ground truth labels. $n_c=3$ corresponds to the number of classes, and $n_{s}$ represents the number of samples in a batch.

During testing, human-firearm bounding box pairs are formed by automatically detecting them using the firearm and human detectors.



\subsection{Preserving Locality of Objects}
\label{sect:local-EHFPL}


{To enhance the FEB's ability to capture locality-aware features, we introduce a decoder network whose output is a saliency-driven locality mask for each object within the PBB.} To accomplish this, the output of the FEB is passed through a series of deconvolution and upsampling layers, which resize the feature map to match the original PBB dimensions. The resulting feature map is then fed into a sigmoid layer, which produces the object presence probability at each pixel. This branch's output consists of three channels with the same spatial dimensions as the original PBB.

To create the ground truth for this branch, we generate masks using a Gaussian probability distribution function centered at the midpoint of the object's bounding box. A diagonal covariance matrix is employed, with variances in both directions proportional to the object's width and height. If an object is absent within a PBB, its corresponding mask is set as an array of all zeros. The three masks for the possible objects (gun, rifle, human) are concatenated to form an image, denoted as $G_{\text{map}}$, which serves as the ground truth for the saliency-driven locality-preserving branch, as illustrated in Fig. \ref{fig:main}. The output of this branch, denoted as $P_{\text{map}}$, is compared to the ground truth using the Frobenius norm.
\begin{equation}
\label{eq:ploss}
\centering
\mathcal{L}_p(P_{map}, G_{map}) = \sum_{i=1}^{n_{s}} ||G_{map} - P_{map}||_F.
\end{equation} 
The decoder is trained to minimize  $\mathcal{L}_p$ loss while the feature encoding block is simultaneously trained to minimize both $\mathcal{L}_p$ and $\mathcal{L}_c$. 


\vspace{-5mm}
\subsection{Overall Loss Function}
During inference we only use the output of PBB classification module to determine firearm carriers. Whole network is  end-to-end trained by minimizing :
\begin{equation}
   \min \left( \mathcal{L}_c+ \lambda \mathcal{L}_p\right ),
\end{equation}
where $\mathcal{L}_c$ is the classification loss, $\mathcal{L}_p$ is the locality preserving loss  and $\lambda$ is the hyper-parameter assigning relative importance to the term. 
Thus the proposed FEB becomes aware of both the classification as well as the object locality and appearance.


\vspace{-5mm}
\subsection{Maxout Detections}
\label{sect:mxout-EHFPL}
Assuming that a firearm shall be carried  by a single person, if a firearm bounding box $b_o$ appears in multiple PBBs we pick the one that is assigned maximum probability by the PBB Classification Network. 
Rest are assigned non-interaction labels. This stage of the architecture is only used during inference.

\vspace{-3mm}
\section{Experiments and Results}
\label{sec:experiments}


The proposed algorithm is extensively evaluated on two datasets: the newly introduced \textit{Localization of Firearm Carriers} (LFC) dataset, and a subset of the existing dataset called `Humans Interacting with Common Objects' (HICO-DET) \cite{HICO_DET}. The performance of our algorithm is compared against resent HOI methods, including Transferable Interactiveness Network (TIN) \cite{li2019transferable}, Instance Centric Attention Network (iCAN) \cite{gao2018ican}, and Parallel Point Detection and Matching (PPDM) \cite{liao2020ppdm}.
\subsection{Localization of Firearm Carriers (LFC) Dataset}

To address the problem of localizing firearm carriers, we introduce a comprehensive dataset dubbed as \textit{Localization of Firearm Carriers} (LFC). This dataset consists of 3,128 images, each depicting at least one human interacting with a firearm in various scenarios. For the LFC dataset, we gather 2,218 images from the OAOD dataset \cite{oaod}, while the remaining images are collected from the internet. During the collection process, we utilize search keywords such as ``human firearms'', ``person carrying firearms/gun/rifle'', and ``person holding firearms/gun/rifle'' to ensure a diverse range of images. To ensure data quality, we carefully curate the dataset by removing irrelevant images, duplicates, and animations.

For each image in the LFC dataset, we manually annotate the bounding boxes for both the firearms and humans. These annotated bounding boxes are then paired together to create Paired Bounding Boxes (PBBs). Each PBB is manually labeled to indicate whether it represents a valid or invalid interaction. In Figure \ref{fig:teaser}, we provide sample images from the LFC dataset to illustrate its content. {On the  LFC dataset, the Average Precision (AP) of rifle detection is 89.8\% and AP of gun detection is 87.9\% using OAOD, and  AP of person detection is 90.7\% (see Sec. \ref{sect:det-EHFPL} for more details). The LFC dataset has been made publicly available \footnote{\textcolor{blue}{Link to LFC Dataset: http://bit.ly/43E9fKd }}.}

The LFC dataset presents several challenges for the problem of Human Firearm Interaction due to varying overlaps between humans and firearms, cluttered backgrounds, viewpoint variations, and complex interaction scenarios. By making the LFC dataset publicly available to the research community, we aim to facilitate advancements in human-object interaction research. The dataset comprises 5.2k training instances (cropped human-object pairs) and 1.3k testing instances, providing a substantial resource for conducting thorough evaluations and developing innovative approaches.

\subsection{HICO-DET Dataset}
\label{sec:HDdataset}

{The HICO-DET benchmark dataset \cite{HICO_DET} serves as a crucial resource for evaluating human-object interactions. It encompasses 80 object classes and 600 Human Object Interactions (HOI) categories. The dataset comprises 47.7K images, with over 150K annotated human-object pairs. For our experiments, we focus on HOI categories that involve humans holding an object. The selected 60 objects include a diverse range including: bicycle, bird, bottle, cat, chair, cow, dog, horse, motorcycle, person, potted plant, sheep, apple, backpack, banana, baseball bat, book, baseball glove, bowl, broccoli, cake, carrot, cellphone, clock, cup, donut, fork, frisbee, hairdryer, handbag, hot dog, keyboard, kite, knife, laptop, mouse, orange, oven, pizza, refrigerator, remote, sandwich, scissors, skateboard, skis, snowboard, spoon, sports ball, stop sign, suitcase, surfboard, teddy bear, tennis racket, tie, toaster, toothbrush, umbrella, vase, wine glass, and zebra.}

The dataset provides 13,963 training images and 3,748 test images for a total of 60 HOI categories associated with these 60 objects. The training set contains over 35K instances with hold/not-hold labels, while the testing set consists of more than 9.7K instances. Notably, it is important to highlight that no existing HOI dataset includes interactions involving firearms, guns, or rifles. Thus, our work is the first, to the best of our knowledge, to address this specific problem and introduce firearm-related interactions to the research community.

\subsection{Evaluation Measures}

To ensure consistency with the state-of-the-art in Human Object Interaction (HOI) evaluation, our evaluation setup aligns with the approach used by Chao et al. \cite{HICO_DET}. We compute the Average Precision (AP) for the ``hold'' category as follows. First, we utilize the trained classifier to estimate the probability of holding for each pair of detected humans and firearms. Then, we calculate the detection score by multiplying the object detection probability with the association probability, following the methodology outlined in \cite{gao2018ican}. The detections (pairs) are sorted based on their computed scores.

For each pair, we evaluate the Intersection over Union ($IoU$) between the detected humans and the ground truth humans, as well as between the detected firearms and the ground truth firearms. A pair is considered a true positive if both $IoU$ values are greater than 0.5. We assign true positive or false positive labels to each pair based on the $IoU$ threshold. Using these counts of true positives and false positives, we calculate precision, recall, and mean Average Precision (mAP) as evaluation metrics for the ``hold'' category.

\subsection{Baseline Experiments on Carried/Not-carried Firearms}

A firearm is considered ``carried'' if any person in the image is carrying the detected firearm; otherwise, the firearm is categorized as ``not carried''. In the HiFD baseline (Section \ref{sect:HiFD}), we classify a detected firearm as ``carried'' if hand keypoints are detected within the firearm bounding box with a probability greater than or equal to 0.30. To retrieve hand keypoints inside firearm bounding boxes, we utilize Simon et al.'s method \cite{simon2017hand} in combination with the firearm detections from OAOD \cite{oaod}.
In the HCFD baseline (Section \ref{sect:HCFD}), human body poses are estimated using OpenPose \cite{cao2018openpose}. A firearm is classified as ``carried'' if at least two hand keypoints are found within the firearm bounding box.

Table \ref{tab:clsgroundtruth} presents the accuracy of these baseline experiments. HiFD performs well in classifying carried/not-carried guns; however, its accuracy is lower for the rifle class. Conversely, HCFD shows reasonable performance in rifle classification as carried or not-carried, but its accuracy is degraded for guns.

For the evaluation of the third baseline OHFD (Section \ref{sect:OHFD}), humans were detected using a pre-trained Faster RCNN model with a ResNet-101 backbone, while OAOD was used for firearm detection. In OHFD, a firearm is classified as ``carried'' if the Intersection over Union (IoU) between the firearm bounding box and the person bounding box is greater than or equal to 0.50. The results obtained from HiFD, HCFD, and OHFD serve as baselines for comparison.

\begin{table}[t]
\centering
\small
\renewcommand{\arraystretch}{1.1}
\tabcolsep=4.5pt\relax
\caption{\textsc{Firearm classification  as carried or not carried  accuracy (\%) comparison.}}
\begin{tabular}{c|c|ccc} 
\hline
\textbf{\begin{tabular}[c]{@{}c@{}} Classifiers\end{tabular}} & \textbf{\begin{tabular}[c]{@{}c@{}}Carrier\\ Identification\end{tabular}} & \textbf{Gun} & \textbf{Rifle} & \textbf{Overall} \\ [0.5ex] 
\hline
\text{\begin{tabular}[c]{@{}c@{}} HiFD \\(Baseline) \end{tabular}} & No  & 71.9 & 37.5 & 49.2 \\ \hline
\text{\begin{tabular}[c]{@{}c@{}} HCFD \\(Baseline) \end{tabular}} & No  & 51.3 & 76.4 & 66.6 \\ \hline
\text{\begin{tabular}[c]{@{}c@{}} OHFD \\(Baseline) \end{tabular}} & Yes & \text{84.1} & 86.8 & 85.9 \\ \hline
\text{\begin{tabular}[c]{@{}c@{}} HFPL \end{tabular}} & Yes & 83.9 & \text{88.4} & \text{87.5} \\ \hline
\textbf{\begin{tabular}[c]{@{}c@{}} A-HFPL \end{tabular}} & Yes & \textit{89.5} & \textit{94.3} & \textit{92.0} \\ \hline
\textbf{\begin{tabular}[c]{@{}c@{}} E-HFPL \end{tabular}} & Yes & \textbf{91.5} & \textbf{94.3} & \textbf{93.25} \\ \hline
\end{tabular}
\label{tab:clsgroundtruth}
\end{table}


\begin{table*}[t]
\centering
\renewcommand{\arraystretch}{1.2}
\tabcolsep=10.5pt\relax
\caption{\small\textsc{Performance comparison of different variants of the proposed algorithm, including HFPL (baseline), A-HFPL (attention), and E-HFPL (saliency driven).
\textbf{$AP_{Rhold}$} is average precision of Rifle, \textbf{$AP_{Ghold}$} is average precision of Gun and \textbf{$AP_{hold}$} is overall average precision.}}
\begin{tabular}{c|c|ccc|ccc} 
\hline
  & \textbf{Color Space} & \multicolumn{3}{c|}{\textbf{without Max-Out Detection}} & \multicolumn{3}{c}{\textbf{with Max-Out Detection}} \\ \hline
 
\textbf{Methods} & - &\textbf{$AP_{Ghold}$} & \textbf{$AP_{Rhold}$} & \textbf{$AP_{hold}$} & \textbf{$AP_{Ghold}$} & \textbf{$AP_{Rhold}$} & \textbf{$AP_{hold}$} \\ \hline





\begin{tabular}[c]{@{}c@{}} HFPL \end{tabular}   & GrayScale  &  59.2  &  63.6   & 61.4  & 62.9  &  65.5  &  64.2   \\ \hline
\begin{tabular}[c]{@{}c@{}} A-HFPL \end{tabular} & GrayScale  &  68.5  &  74.4   & 71.6  & 69.7  &  74.6  &  72.1   \\ \hline
\begin{tabular}[c]{@{}c@{}} E-HFPL \end{tabular} & GrayScale  &  70.2  &  76.0   & 73.1  & 71.4  &  75.8  &  73.6   \\ \hline

\begin{tabular}[c]{@{}c@{}} HFPL \end{tabular}   & RGB  &         60.8   &          65.1   &          63.1   &         63.1   &         67.1   &         65.2 \\ \hline
\begin{tabular}[c]{@{}c@{}} A-HFPL \end{tabular} & RGB  &         71.9   &          77.2   &          74.1   &         73.8   &         77.8   &         75.7 \\ \hline
\begin{tabular}[c]{@{}c@{}} E-HFPL \end{tabular} & RGB  & \textit{72.5}  &  \textit{78.5}  &  \textit{75.7}  & \textit{74.3}  & \textit{79.1}  & \textit{76.4} \\ \hline

\begin{tabular}[c]{@{}c@{}} HFPL \end{tabular}    & YCbCr  &         61.9   &          65.7   &          63.8   &         64.6   &         68.1   &         66.3 \\ \hline
\begin{tabular}[c]{@{}c@{}} A-HFPL \end{tabular}  & YCbCr  &         72.2   &          77.6   &          74.9   &         73.9   &         78.4   &         76.1 \\ \hline
\begin{tabular}[c]{@{}c@{}} E-HFPL \end{tabular}  & YCbCr  & \textbf{73.8}  &  \textbf{79.4}  &  \textbf{76.7}  & \textbf{74.7}  & \textbf{80.8}  & \textbf{77.8} \\ \hline
\end{tabular}
\label{tab:attentionours}
\end{table*}

\subsection{Attention-HFPL (A-HFPL)}
\label{sect:AFPL}
To analyze the contribution of each block in our approach, we have modified the existing base method described in \cite{iciploc} by incorporating an attention mechanism. We evaluate the performance of this modified architecture, which aims to address the limitations of the base algorithm discussed in Section \ref{sect:intro}. The attention mechanism focuses solely on the queried firearm and person bounding boxes, enabling the architecture to distinguish the actual objects of interest from background firearms and persons.

The pipeline of the attention-based Human Firearm Pair Localization (HFPL) is similar to the one used in \cite{iciploc}, with the inclusion of attention channels. The attention mechanism utilizes the bounding box information of humans and objects to generate two masks. Each mask is represented as a matrix of zeros with dimensions equal to the input Paired Bounding Boxes (PBBs). For humans, the mask assigns a value of one to pixels located inside the human bounding box, while for firearms, the mask assigns a value of one to pixels within the object bounding box. These masks are then concatenated with the cropped instances, resulting in a 5-dimensional input representation.
\subsection{Comparison of PBB Localization \& Classification}
In these experiments, we employ multi-sized Paired Bounding Boxes (PBBs) for training the classification network. Additionally, we explore different color spaces during the experiment. For the training of HFPL (no-attention), A-HFPL (attention-based), and E-HFPL (attention-based with modified architecture) with Adaptive Average Pooling (AAP), we set the learning rate to 1.0 $\times$ $e^{-5}$, use a batch size of 1, and apply dropout at a rate of 50

To train the models, we utilize stochastic gradient descent (SGD) with a momentum of 0.9. The training process is conducted for 30 epochs. For both training and evaluation, we make use of the LFC dataset. The training set comprises 2,400 images, with a total of 2,831 positive instances and 2,280 negative instances of human-firearm pairs that have been annotated. The testing set consists of 700 images, with 800 positive instances and 467 negative instances.

To ensure consistency in the input sizes, the multi-sized PBBs are resized in such a way that the shorter dimension corresponds to 480 pixels, while the longer side adjusts accordingly to maintain the aspect ratio.

\subsubsection{HFPL} To evaluate the effectiveness of localizing the human-firearm pair in the image we train HFPL on the LFC dataset. As in HFPL{\cite{iciploc}}, we use ground-truth PBBs to train the classification module, which consists of VGG-16 based model, pre-trained on ImageNet.


\subsubsection{A-HFPL} For this experiment, channel-wise attention is concatenated with the generated PBBs and forwarded to the network. For RGB and YCbCr, five channels of input are generated that incorporate \textit{Attention Channels} for firearm and human respectively as its last two channels. In the case of gray-scale images, concatenating \textit{Attention Channels} results in three-channel input.  The improvement in mAP by A-HFPL over HFPL can be seen in Table \ref{tab:attentionours}.


\subsubsection{E-HFPL} In this experiment, we introduce modifications to the attention mechanism as described in Section \ref{sect:atten-EHFPL}. Firearms and humans are detected in the training dataset, and PBBs are formed using detections with object probabilities greater than 0.5. To determine the inclusion of a detected PBB as a positive example, we consider the IOU (Intersection over Union) between the person and firearm present in the PBB and the ground truth person and firearm. If the IOU value is greater than 0.5, the PBB is included as a positive example based on the interaction label of their corresponding ground truth human and firearm. This approach enables the model to learn and rectify mistakes made by object detection models to some extent.

The incorporation of attention channels allows the convolutional neural networks (CNNs) to focus on more salient regions, facilitating more accurate decision-making regarding interactiveness. By directing the model's attention to relevant regions, the E-HFPL architecture aims to improve the overall performance of the system.

\begin{figure*}[t]
    \begin{center}
    \includegraphics[width=1.0\linewidth]{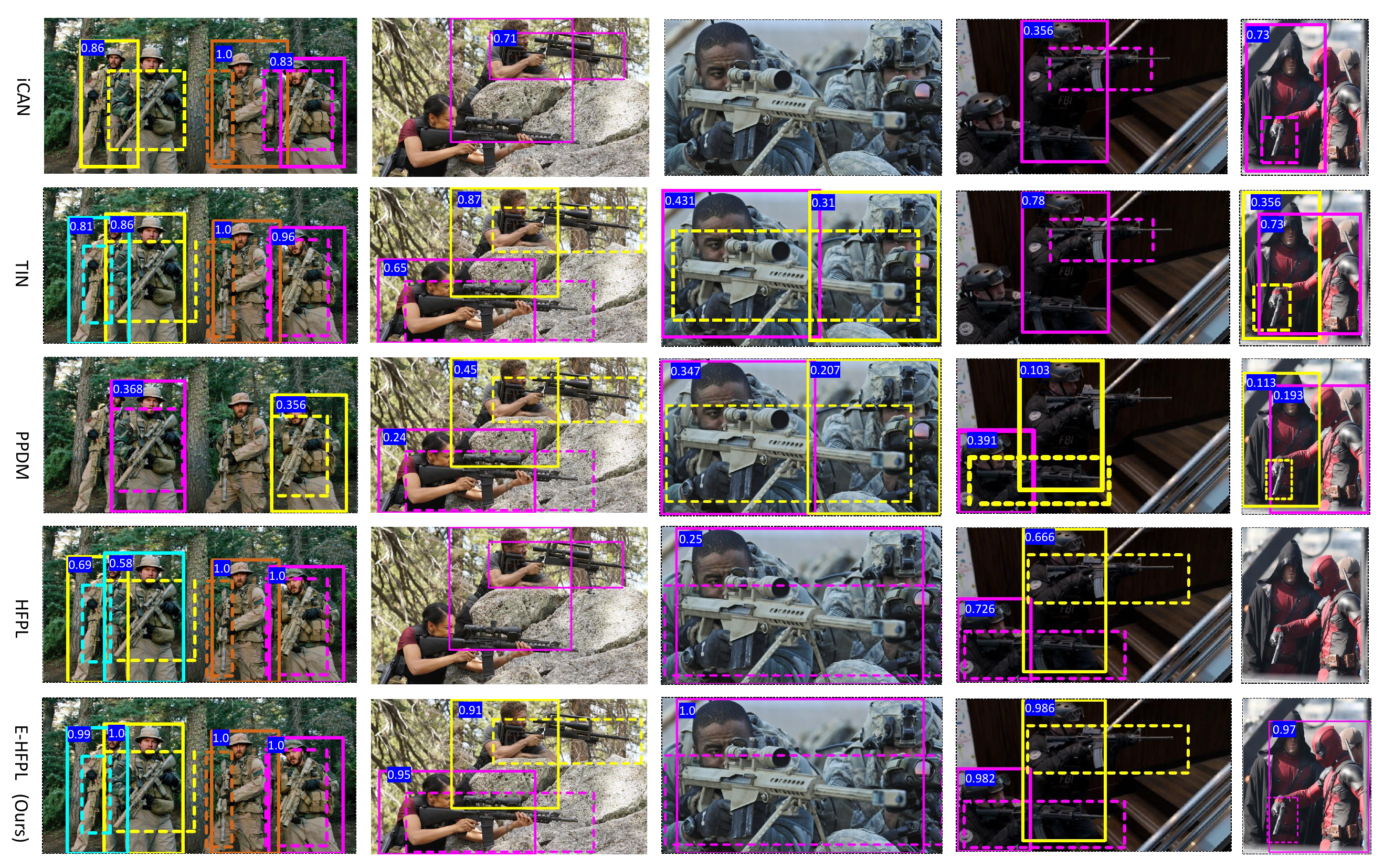}
    \caption{Visual Comparison between different approaches: Our method identifies human-firearm pair with high scores as compared to other baseline and HOI methods.
    There is miss-association in other methods as well.
    Strong lines (human detection), and dotted lines (the same color associated with firearms and  humans).}
    \label{fig:visual}
    \end{center}
\end{figure*}

The comparisons of $AP_{hold}$ over different color spaces for E-HFPL, A-HFPL, and HFPL are shown in Table \ref{tab:attentionours}. 
The classification accuracy shown in Table \ref{tab:clsgroundtruth} is computed on the testing set using ground truth bounding boxes of humans and firearms.
Fig \ref{fig:visual} shows the output score and strength of our proposed method E-HFPL. 

\subsubsection{Inference Details} During inference, we use Faster RCNN (ResNet-101 \& MS-COCO) \cite{ren2015faster} for human prediction and OAOD \cite{oaod} for firearm prediction. PBBs are generated on these predictions and evaluated on the described metrics.

\begin{table}[b]
\centering
\renewcommand{\arraystretch}{1.2}
\tabcolsep=2.0pt\relax
\caption{\small\textsc{Comparison of our proposed algorithm with State of the art HOI-based algorithms on LFC Dataset and HICO-DET Dataset.
HICO-DET(S) shows the subset of the HICO-DET dataset.
}}.
\begin{tabular}{|c|c|c|c|c|} 
\hline
 
\textbf{Methods} & \textit{iCAN \cite{gao2018ican}} & \textit{TIN \cite{li2019transferable}} & \textit{PPDM \cite{liao2020ppdm}} & \textit{E-HFPL} \\ \hline
\multicolumn{5}{|c|}{\textbf{LFC}}  \\ \hline

$AP_{hold}$ & 68.1 & \underline{76.1} & \text{63.3} & \textbf{77.8}    \\ \hline


\multicolumn{5}{|c|}{\textbf{HICO-DET (S)}}  \\ \hline
 

$AP_{hold}$ & 33.5 & 35.1 & \underline{38.8} & \textbf{40.2}    \\ \hline

\end{tabular}
\label{tab:comparisonhico}
\end{table}

\subsection{Comparison with HOI-based Methods}

We conducted a comprehensive comparison of our method with several existing HOI-based algorithms. The experiments were performed on our proposed dataset as well as a subset of the HICO-DET dataset, as discussed in Section \ref{sec:HDdataset}. To ensure a fair comparison, all methods were trained exclusively for the hold category from HICO-DET, using the same training set.

Table \ref{tab:comparisonhico} presents the comparison of E-HFPL (Ours) with other HOI-based algorithms. Our algorithm demonstrates a significant advantage over the compared methods, primarily due to its customized design specifically tailored for the hold category, while the compared HOI methods focus on addressing inter-class variations. We compare our method with iCAN \cite{gao2018ican}, TIN \cite{li2019transferable}, and PPDM \cite{liao2020ppdm}. In all experiments, the hyperparameters were set as defined in the respective HOI methods. However, the compared HOI methods fail to achieve improved results on our proposed dataset. The simplicity of our proposed method, relying on attention and saliency-driven locality-preserving loss, allows it to outperform the compared HOI methods.

For the training of iCAN, we considered four interactions: gun and rifle interaction and non-interaction, along with three objects: human, gun, and rifle, as specified in the LFC dataset. We followed the recommended settings and used detectron's Faster RCNN \cite{wu2019detectron2} for human detection and OAOD \cite{oaod} for firearm detection. The learning rate was set to 0.001, and the batch size was 1 with 180K iterations.

The PPDM method was trained using similar objects: human, gun, and rifle, with interactions among all objects. It is an end-to-end system that predicts both detection and interaction in the same pipeline. We trained PPDM with a learning rate of 4.5 $\times$ $e^{-4}$, a batch size of 12, and 30K iterations.

Figure \ref{fig:visual} provides visual results comparing the HOI-based methods with our method, E-HFPL. It is important to note that our method handles negatives differently than iCAN \cite{gao2018ican}. While iCAN relies on thresholding based on $IoU$ to determine positives and negatives, we use a different mechanism based on the interactiveness component, distinguishing between interaction and non-interaction. Additionally, negatives are incorporated when the object detection algorithm fails to precisely detect the object, indicated by $IoU <$ 0.50.

\subsection{Discussions and Analysis}

Our experiments demonstrate that E-HFPL surpasses other baselines and state-of-the-art methods in terms of accuracy. In the task of identifying Firearm Carriers (Table ~\ref{tab:clsgroundtruth}), E-HFPL achieves an overall accuracy that is approximately 5.7 points higher than HFPL. The integration of the locality-preserving layer and attention channels leads to state-of-the-art accuracy for both Gun and Rifle carrier identification, highlighting the effectiveness of E-HFPL.

In the context of Human-Object Interaction, E-HFPL outperforms existing methods such as iCAN, TIN, and PPDM. On the LFC dataset, E-HFPL achieves an improvement of 1.6 points compared to the previous state-of-the-art. Even when evaluated on the subset of the HICO-DET dataset, which consists of 60 objects held by humans, our proposed method exhibits significant improvements in mAP.

Additionally, we conducted detailed experiments by exploring different color spaces in combination with attention channels (Table \ref{tab:attentionours}). Grayscale images, which lack color information, are less informative than RGB images. The RGB color space allows the model to better leverage color attributes compared to grayscale. Among the color spaces evaluated, YCbCr proves to be superior to RGB as it effectively separates the luminance property from the other planes' intensity. This separation enables the model to learn features specifically from the luminance components of the images.

 








\begin{figure}[t]
\centering
\includegraphics[width=8.5cm]{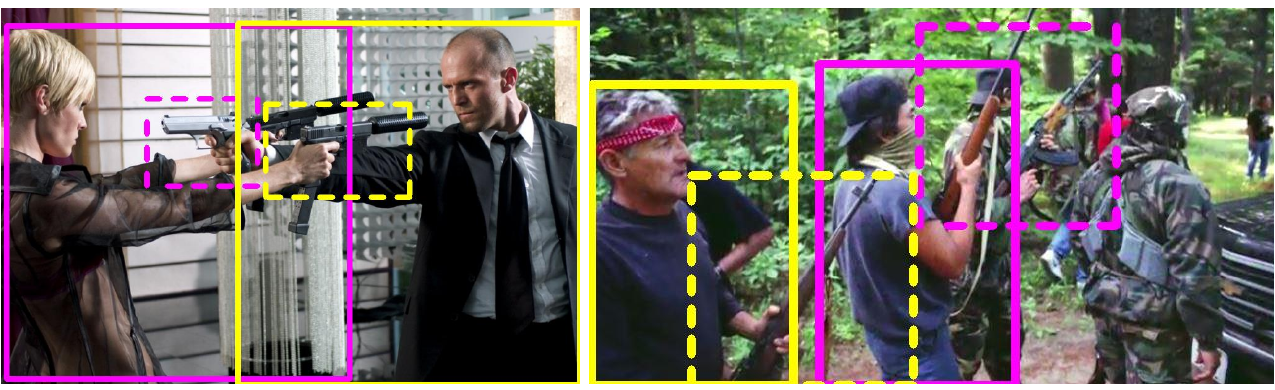}
\caption{Failure cases: Objects too far from the carrier in the presence of another human may result in failure of association. Likewise for overlapping firearms and occluded humans.
}
\label{fig:failure}
\end{figure}



\subsection{Failure Cases} 

While our model achieves impressive results, there are certain scenarios where its performance is impacted. These failure cases primarily occur when there are overlapping humans and firearms, as well as increased distance and occlusion between them. In such complex situations, accurately predicting the association between a human and the respective firearm becomes challenging for our model.



\section{Conclusion and Future Directions}
\label{sec:majhead}


In conclusion, this paper introduces a novel attention-based mechanism for accurately localizing firearm carriers in complex scenes. This solution addresses the need for effective surveillance in security management systems. By leveraging information from humans and firearms and integrating spatially salient regions through attention channels, our approach improves the classification accuracy of localized human-firearm pairs. Additionally, the saliency-driven locality-preserving branch ensures that the extracted features are informative for pair classification.

We investigate the impact of attention channels and employ adaptive average pooling to handle multi-size paired bounding boxes effectively. By leveraging all these components, our method achieves state-of-the-art results, with an impressive 77.8\% $AP_{hold}$ on the firearms test set. Furthermore, when compared with HOI-based methods, our simple yet effective approach outperforms in terms of the ``hold'' or ``not hold'' classification.


In summary, our attention-based mechanism, combined with the utilization of firearms and human information, spatial saliency, and adaptive average pooling, significantly advances the task of localizing firearm carriers and achieves remarkable performance in challenging scenarios. This research contributes to enhancing surveillance capabilities and holds promise for improving security management systems.

\begin{thebibliography}{10}
\providecommand{\url}[1]{#1}
\csname url@samestyle\endcsname
\providecommand{\newblock}{\relax}
\providecommand{\bibinfo}[2]{#2}
\providecommand{\BIBentrySTDinterwordspacing}{\spaceskip=0pt\relax}
\providecommand{\BIBentryALTinterwordstretchfactor}{4}
\providecommand{\BIBentryALTinterwordspacing}{\spaceskip=\fontdimen2\font plus
\BIBentryALTinterwordstretchfactor\fontdimen3\font minus
  \fontdimen4\font\relax}
\providecommand{\BIBforeignlanguage}[2]{{%
\expandafter\ifx\csname l@#1\endcsname\relax
\typeout{** WARNING: IEEEtran.bst: No hyphenation pattern has been}%
\typeout{** loaded for the language `#1'. Using the pattern for}%
\typeout{** the default language instead.}%
\else
\language=\csname l@#1\endcsname
\fi
#2}}
\providecommand{\BIBdecl}{\relax}
\BIBdecl

\bibitem{TCSS1}
P.~Lu, D.~Chen, Y.~Li, X.~Wang, and S.~Yu, ``Agent-based model of mass campus
  shooting: Comparing hiding and moving of civilians,'' \emph{IEEE Transactions
  on Computational Social Systems}, pp. 1--10, 2022.

\bibitem{newyear2020}
``In a deadly start to 2020, gunfire erupts across america - the new york
  times,'' \url{https://www.nytimes.com/2020/01/01/us/new-year-shootings.html},
  (Accessed on 01/11/2021).

\bibitem{Americas23:bbc}
``America's gun culture in charts - bbc news,''
  \url{https://www.bbc.com/news/world-us-canada-41488081}, (Accessed on
  01/11/2021).

\bibitem{amnesty}
``Gun violence key facts amnesty international,''
  \url{https://www.amnesty.org/en/what-we-do/arms-control/gun-violence/},
  (Accessed on 01/11/2021).

\bibitem{howard2013suspiciousness}
C.~J. Howard, T.~Troscianko, I.~D. Gilchrist, A.~Behera, and D.~C. Hogg,
  ``Suspiciousness perception in dynamic scenes: a comparison of cctv operators
  and novices,'' \emph{Front. in Human Neuro.}, vol.~7, p. 441, 2013.

\bibitem{TCSS2}
T.~Lin and H.~Pham, ``Modeling security surveillance systems with state
  dependent inspection-maintenance strategy,'' \emph{IEEE Transactions on
  Computational Social Systems}, pp. 1--12, 2022.

\bibitem{TCSS4}
S.~Madichetty and S.~M, ``A neural-based approach for detecting the situational
  information from twitter during disaster,'' \emph{IEEE Transactions on
  Computational Social Systems}, vol.~8, no.~4, pp. 870--880, 2021.

\bibitem{TCSS3}
B.~Zhou, L.~Chen, S.~Zhao, S.~Li, Z.~Zheng, and G.~Pan, ``Unsupervised domain
  adaptation for crime risk prediction across cities,'' \emph{IEEE Transactions
  on Computational Social Systems}, pp. 1--11, 2022.

\bibitem{oaod}
J.~Iqbal, M.~A. Munir, A.~Mahmood, A.~R. Ali, and M.~Ali, ``Orientation aware
  object detection with application to firearms,'' \emph{arXiv preprint
  arXiv:1904.10032}, 2019.

\bibitem{iciploc}
A.~{Basit}, M.~A. {Munir}, M.~{Ali}, N.~{Werghi}, and A.~{Mahmood},
  ``Localizing firearm carriers by identifying human-object pairs,'' in
  \emph{IEEE International Conference on Image Processing}, 2020, pp.
  2031--2035.

\bibitem{ren2015faster}
S.~Ren, K.~He, R.~Girshick, and J.~Sun, ``Faster r-cnn: Towards real-time
  object detection with region proposal networks,'' in \emph{Advances in neural
  information processing systems}, 2015, pp. 91--99.

\bibitem{redmon2018yolov3}
J.~Redmon and A.~Farhadi, ``Yolov3: An incremental improvement,'' \emph{arXiv
  preprint arXiv:1804.02767}, 2018.

\bibitem{liu2016ssd}
W.~Liu, D.~Anguelov, D.~Erhan, C.~Szegedy, S.~Reed, C.-Y. Fu, and A.~C. Berg,
  ``Ssd: Single shot multibox detector,'' in \emph{European conference on
  computer vision}.\hskip 1em plus 0.5em minus 0.4em\relax Springer, 2016, pp.
  21--37.

\bibitem{lin2017feature}
T.-Y. Lin, P.~Doll{\'a}r, R.~Girshick, K.~He, B.~Hariharan, and S.~Belongie,
  ``Feature pyramid networks for object detection,'' in \emph{Proceedings of
  the IEEE conference on computer vision and pattern recognition}, 2017, pp.
  2117--2125.

\bibitem{tian2019fcos}
Z.~Tian, C.~Shen, H.~Chen, and T.~He, ``Fcos: Fully convolutional one-stage
  object detection,'' in \emph{Proceedings of the IEEE International Conference
  on Computer Vision}, 2019, pp. 9627--9636.

\bibitem{duan2019centernet}
K.~Duan, S.~Bai, L.~Xie, H.~Qi, Q.~Huang, and Q.~Tian, ``Centernet: Keypoint
  triplets for object detection,'' in \emph{Proceedings of the IEEE
  International Conference on Computer Vision}, 2019, pp. 6569--6578.

\bibitem{singh2018analysis}
B.~Singh and L.~S. Davis, ``An analysis of scale invariance in object detection
  snip,'' in \emph{Proceedings of the IEEE conference on computer vision and
  pattern recognition}, 2018, pp. 3578--3587.

\bibitem{redmon2017yolo9000}
J.~Redmon and A.~Farhadi, ``Yolo9000: better, faster, stronger,'' in
  \emph{Proceedings of the IEEE conference on computer vision and pattern
  recognition}, 2017, pp. 7263--7271.

\bibitem{redmon2016you}
J.~Redmon, S.~Divvala, R.~Girshick, and A.~Farhadi, ``You only look once:
  Unified, real-time object detection,'' in \emph{IEEE conference on computer
  vision and pattern recognition}, 2016, pp. 779--788.

\bibitem{fu2017dssd}
C.~Fu, W.~Liu, A.~Ranga, A.~Tyagi, and A.~Berg, ``Dssd: Deconvolutional single
  shot detector,'' \emph{arXiv preprint arXiv:1701.06659}, 2017.

\bibitem{iqbal2019orientation}
J.~Iqbal, M.~A. Munir, A.~Mahmood, A.~R. Ali, and M.~Ali, ``Orientation aware
  object detection with application to firearms,'' \emph{arXiv preprint
  arXiv:1904.10032}, vol.~22, 2019.

\bibitem{olmos2018automatic}
R.~Olmos, S.~Tabik, and F.~Herrera, ``Automatic handgun detection in videos
  using deep learning,'' \emph{Neurocomputing}, vol. 275, pp. 66--72, 2018.

\bibitem{akcay2018using}
S.~Akcay, M.~E. Kundegorski, C.~G. Willcocks, and T.~P. Breckon, ``Using deep
  convolutional neural network architectures for object classification and
  detection within x-ray baggage security imagery,'' \emph{IEEE transactions on
  information forensics and security}, vol.~13, no.~9, pp. 2203--2215, 2018.

\bibitem{Rogers2017ADL}
T.~W. Rogers, N.~Jaccard, and L.~D. Griffin, ``A deep learning framework for
  the automated inspection of complex dual-energy x-ray cargo imagery,'' in
  \emph{Anomaly Detection and Imaging with X-Rays (ADIX) II}, vol. 10187.\hskip
  1em plus 0.5em minus 0.4em\relax SPIE, 2017, pp. 106--117.

\bibitem{wang2019deep}
T.~Wang, R.~M. Anwer, M.~H. Khan, F.~S. Khan, Y.~Pang, L.~Shao, and
  J.~Laaksonen, ``Deep contextual attention for human-object interaction
  detection,'' in \emph{Proceedings of the IEEE International Conference on
  Computer Vision}, 2019, pp. 5694--5702.

\bibitem{gkioxari2018detecting}
G.~Gkioxari, R.~Girshick, P.~Doll{\'a}r, and K.~He, ``Detecting and recognizing
  human-object interactions,'' in \emph{Proceedings of the IEEE Conference on
  Computer Vision and Pattern Recognition}, 2018, pp. 8359--8367.

\bibitem{simon2017hand}
T.~Simon, H.~Joo, I.~Matthews, and Y.~Sheikh, ``Hand keypoint detection in
  single images using multiview bootstrapping,'' in \emph{CVPR}, 2017.

\bibitem{cao2018openpose}
Z.~Cao, G.~Hidalgo, T.~Simon, S.-E. Wei, and Y.~Sheikh, ``Openpose: realtime
  multi-person 2d pose estimation using part affinity fields,'' \emph{arXiv
  preprint arXiv:1812.08008}, 2018.

\bibitem{wei2016cpm}
S.-E. Wei, V.~Ramakrishna, T.~Kanade, and Y.~Sheikh, ``Convolutional pose
  machines,'' in \emph{CVPR}, 2016.

\bibitem{li2019crowdpose}
J.~Li, C.~Wang, H.~Zhu, Y.~Mao, H.-S. Fang, and C.~Lu, ``Crowdpose: Efficient
  crowded scenes pose estimation and a new benchmark,'' in \emph{Proceedings of
  the IEEE Conference on Computer Vision and Pattern Recognition}, 2019, pp.
  10\,863--10\,872.

\bibitem{gao2018ican}
C.~Gao, Y.~Zou, and J.-B. Huang, ``ican: Instance-centric attention network for
  human-object interaction detection,'' \emph{arXiv preprint arXiv:1808.10437},
  2018.

\bibitem{li2019transferable}
Y.-L. Li, S.~Zhou, X.~Huang, L.~Xu, Z.~Ma, H.-S. Fang, Y.~Wang, and C.~Lu,
  ``Transferable interactiveness knowledge for human-object interaction
  detection,'' in \emph{Proceedings of the IEEE Conference on Computer Vision
  and Pattern Recognition}, 2019, pp. 3585--3594.

\bibitem{li2020pastanet}
Y.-L. Li, L.~Xu, X.~Liu, X.~Huang, Y.~Xu, S.~Wang, H.-S. Fang, Z.~Ma, M.~Chen,
  and C.~Lu, ``Pastanet: Toward human activity knowledge engine,'' in
  \emph{Proceedings of the IEEE/CVF Conference on Computer Vision and Pattern
  Recognition}, 2020, pp. 382--391.

\bibitem{liao2020ppdm}
Y.~Liao, S.~Liu, F.~Wang, Y.~Chen, C.~Qian, and J.~Feng, ``Ppdm: Parallel point
  detection and matching for real-time human-object interaction detection,'' in
  \emph{Proceedings of the IEEE/CVF Conference on Computer Vision and Pattern
  Recognition}, 2020, pp. 482--490.

\bibitem{lim2023ernet}
J.~Lim, V.~M. Baskaran, J.~M.-Y. Lim, K.~Wong, J.~See, and M.~Tistarelli,
  ``Ernet: An efficient and reliable human-object interaction detection
  network,'' \emph{IEEE Transactions on Image Processing}, vol.~32, pp.
  964--979, 2023.

\bibitem{park2023viplo}
J.~Park, J.-W. Park, and J.-S. Lee, ``Viplo: Vision transformer based
  pose-conditioned self-loop graph for human-object interaction detection,'' in
  \emph{Proceedings of the IEEE/CVF Conference on Computer Vision and Pattern
  Recognition}, 2023, pp. 17\,152--17\,162.

\bibitem{ma2023fgahoi}
S.~Ma, Y.~Wang, S.~Wang, and Y.~Wei, ``{FGAHOI}: Fine-grained anchors for
  human-object interaction detection,'' \emph{preprint arXiv:2301.04019}, 2023.

\bibitem{imagenet_cvpr09}
J.~Deng, W.~Dong, R.~Socher, L.-J. Li, K.~Li, and L.~Fei-Fei, ``{ImageNet: A
  Large-Scale Hierarchical Image Database},'' in \emph{CVPR09}, 2009.

\bibitem{HICO_DET}
Y.-W. Chao, Y.~Liu, X.~Liu, H.~Zeng, and J.~Deng, ``Learning to detect
  human-object interactions,'' \emph{2018 IEEE Winter Conference on
  Applications of Computer Vision (WACV)}, pp. 381--389, 2018.

\bibitem{wu2019detectron2}
Y.~Wu, A.~Kirillov, F.~Massa, W.-Y. Lo, and R.~Girshick, ``Detectron2,''
  \url{https://github.com/facebookresearch/detectron2}, 2019.

\end{thebibliography}


\vspace{-12mm}
\begin{IEEEbiography}[{\includegraphics[width=1in,height=1.25in,clip,keepaspectratio]{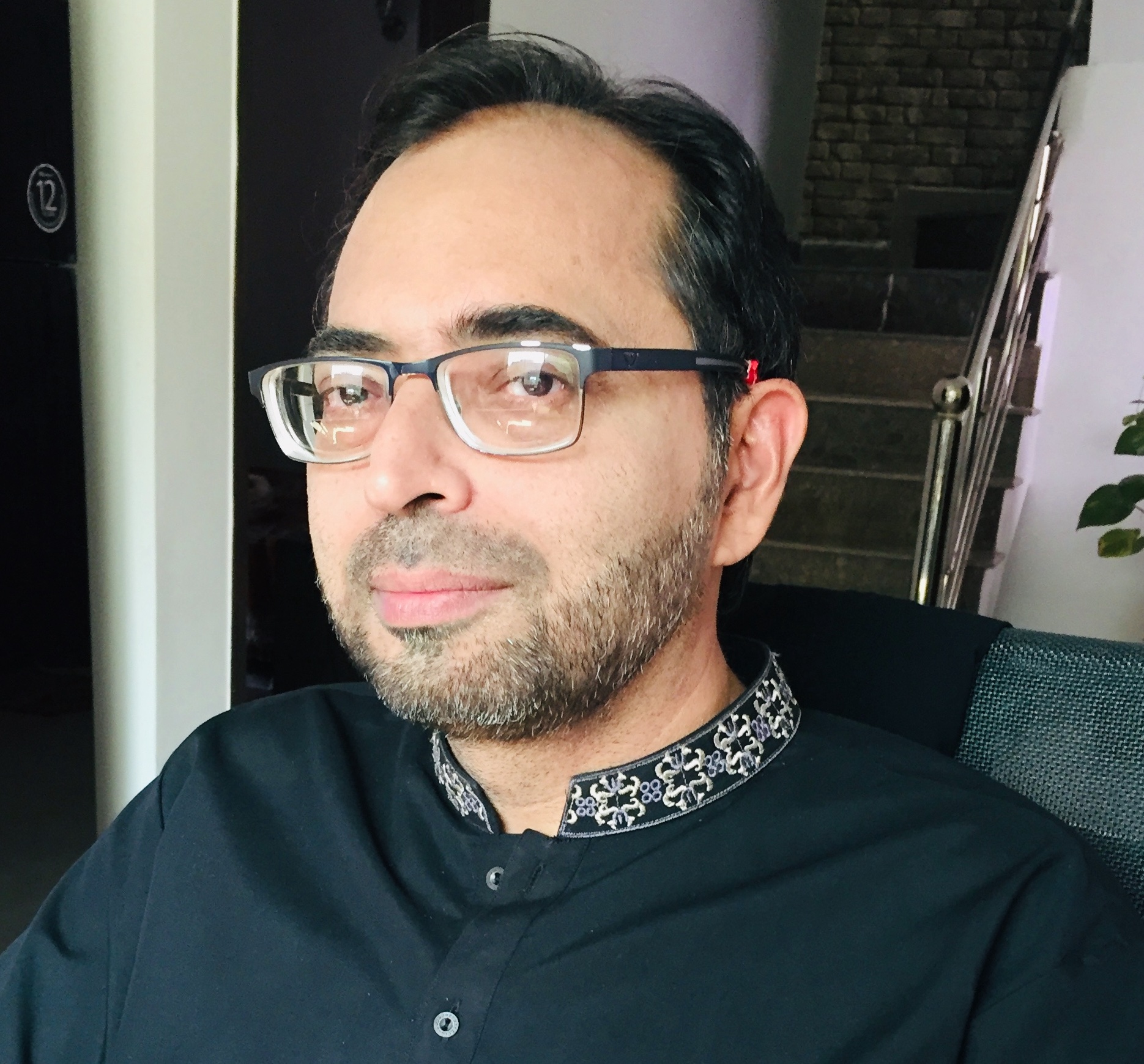}}]{Arif Mahmood} is a Professor in the Department of Computer Science, Dean Faculty of Sciences,  Information Technology University. He is also the Director of the Center for Artificial Intelligence and Robot Vision (CAIRV) at ITU. Previously he worked as a Research Assistant Professor at the University of Western Australia and a PostDoc Researcher at Qatar University. His broad areas of interest include Robot Vision and Machine Learning. More specifically, he is working on moving objects detection in videos, visual object tracking, visual object categorization, nucleus detection, and tissue phenotype in cancer histology images, face detection and facial expression synthesis, action recognition, visual crowd analysis, anomalous event detection,  human body pose estimation, and unsupervised representation learning.
\end{IEEEbiography}
\vspace{-5mm}
\begin{IEEEbiography}[{\includegraphics[width=1in,height=1.25in,clip,keepaspectratio]{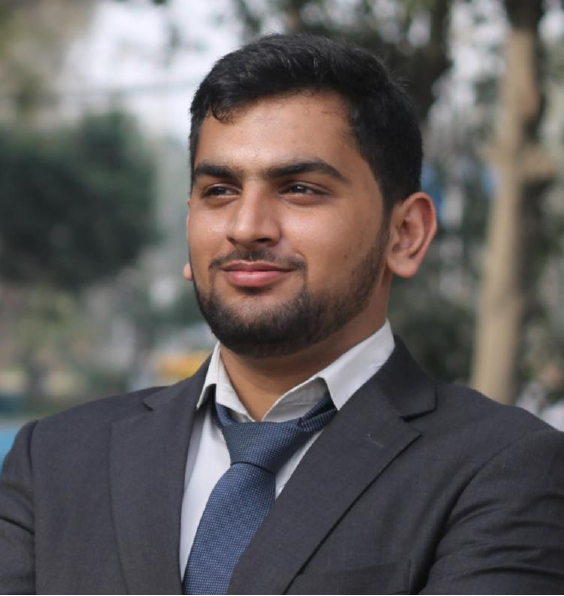}}]{Abdul Basit} received B.Sc degree in Electrical Engineering from the University of Engineering and Technology (UET), Lahore, Pakistan. Currently, he is pursuing M.Sc. in Computer Science from Information Technology University (ITU), Lahore, Pakistan. His area of research is Computer Vision, Deep Learning. He is currently working on detecting Human-Object interactions in violent scenes at the Center for Robot Vision (CRV) at ITU.
\end{IEEEbiography}
\vspace{-10mm}
\begin{IEEEbiography}[{\includegraphics[width=1in,height=1.25in,clip,keepaspectratio]{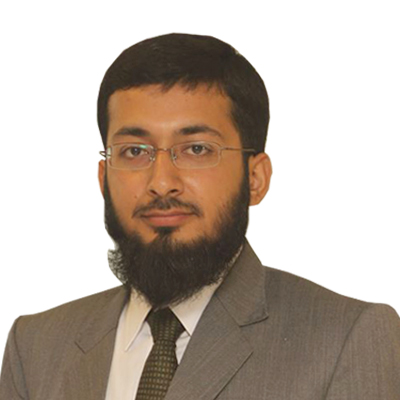}}]{Muhammad Akhtar Munir}
is a Ph.D. research scholar at Intelligent Machines Lab, Information Technology University, Lahore, Pakistan. He received his B.S. and M.S. degrees from COMSATS University Islamabad, Pakistan. His research interests include Object Detection, Domain Adaptation, and Deep Learning. More specifically, he is working on oriented object detection and domain adaptation for object detection.
\end{IEEEbiography}
\vspace{-10mm}
\begin{IEEEbiography}[{\includegraphics[width=1in,height=1.25in,clip,keepaspectratio]{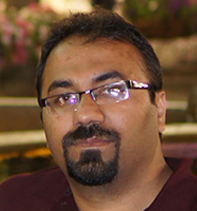}}]{Mohsen Ali} received the M.S. degree in computer science from Lahore University of Management Sciences, Lahore, Pakistan and the Ph.D. degree in computer science from University of Florida, United States. He is an Associate Professor at the Department of Computer Science of Information Technology University. He is a co-founder of the Intelligent Machines Lab at ITU and heads Computer Vision $\&$ Machine Learning Research Group.
His current research interests include Computer Vision and Machine Learning, specifically Deep Learning and Domain Adaptation.
\end{IEEEbiography}
\end{document}